\pdfoutput=1

\documentclass[11pt]{article}

\usepackage[final]{acl}

\usepackage{times}
\usepackage{latexsym}

\usepackage[T1]{fontenc}

\usepackage[utf8]{inputenc}
\usepackage{enumitem}
\usepackage{microtype}

\usepackage{inconsolata}

\usepackage{graphicx}

\usepackage{amsmath,amssymb,amsfonts}
\usepackage{algorithmic}
\usepackage{graphicx}
\usepackage{textcomp}
\usepackage{xcolor}
\usepackage{todonotes}
\usepackage{booktabs}
\usepackage{url}
\usepackage{microtype}
\usepackage{multirow}

\usepackage{arydshln}

\title{Zero-Shot Text-to-Speech for Vietnamese}

\author{Thi Vu, Linh The Nguyen, Dat Quoc Nguyen\\
Movian AI, Vietnam\\
\texttt{\{thivuxy, toank45sphn, datquocnguyen\}@gmail.com}
}

\begin{document}

\maketitle 

\begin{abstract}
This paper introduces PhoAudiobook, a newly curated dataset comprising 941 hours of high-quality audio for Vietnamese text-to-speech. Using PhoAudiobook, we conduct experiments on three leading zero-shot TTS models: VALL-E, VoiceCraft, and XTTS-V2. Our findings demonstrate that PhoAudiobook consistently enhances model performance across various metrics. Moreover, VALL-E and VoiceCraft exhibit superior performance in synthesizing short sentences, highlighting their robustness in handling diverse linguistic contexts. We publicly release PhoAudiobook to facilitate further research and development in Vietnamese text-to-speech.
\end{abstract}

\section{Introduction}
\label{sec:intro}

Text-to-speech (TTS) synthesis has witnessed significant advancements in recent years. 
State-of-the-art TTS systems typically use a cascaded pipeline that consists of an acoustic model and a vocoder, with mel-spectrograms serving as intermediate representations \cite{ren2019fastspeech,li2019neural,shen2018natural}. 
These advanced TTS systems can synthesize high-quality speech for single or multiple speakers \cite{kim2021conditional,liu2022multi}. 

Zero-shot TTS has emerged as a promising approach to overcome the limitations of traditional TTS systems in generalizing to unseen speakers. By leveraging techniques such as speaker adaptation  and speaker encoding, zero-shot TTS aims to synthesize speech for new speakers using only a few seconds of reference audio \cite{chen2019sample,wang2020attentron,9054535,wu22f_interspeech,casanova2022yourtts,xtts}. Recent works have explored the application of language modeling approaches to zero-shot TTS, achieving impressive results. For example, VALL-E \cite{valle} introduces a text-conditioned language model trained on discrete audio codec tokens, enabling TTS to be treated as a conditional codec language modeling task. VoiceCraft \cite{voicecraft} casts both sequence infilling-based speech editing and continuation-based zero-shot TTS as a left-to-right language modeling problem by rearranging audio codec tokens. 

Despite advancements in zero-shot TTS, its application to low-resource languages remains challenging. These languages often lack the large-scale, high-quality datasets needed to train robust TTS models \cite{gutkin-etal-2016-tts,chen19f_interspeech,lux-etal-2022-low,PhamChi_2023,huang-etal-2024-make}. Also, linguistic and phonetic differences between languages introduce additional challenges in adapting existing models to new languages. As a result, the performance of zero-shot TTS systems in low-resource languages is often limited, hindering their practical usability.

In this paper, we focus on advancing zero-shot TTS for Vietnamese. Our contributions are: 
\begin{itemize}[leftmargin=*]
\setlength\itemsep{-1pt}
    \item \textbf{We present PhoAudiobook}, a 941-hour high-quality long-form speech dataset that overcomes the limitations of existing Vietnamese datasets, which usually contain audio samples shorter than 10 seconds. The pipeline to create this dataset can be easily adapted to other languages.
    
    \item \textbf{We conduct a comprehensive experimental study} to evaluate the performance of three state-of-the-art zero-shot TTS models: VALL-E, VoiceCraft, and XTTS-v2~\cite{xtts}. Using a combination of objective and subjective metrics across multiple benchmark datasets, our results demonstrate that XTTS-v2 trained on PhoAudiobook outperforms its counterpart trained on an existing dataset. Additionally, VALL-E and VoiceCraft exhibit robustness in synthesizing varied input lengths.
    
    \item \textbf{We publicly release PhoAudiobook} at \url{https://huggingface.co/datasets/thivux/phoaudiobook} for non-commercial purposes.
\end{itemize}

\section{Dataset}\label{sec:datasets}

\subsection{PhoAudiobook}\label{subsec:phoaudiobook}

\begin{figure}
    \centering
    \includegraphics[width=1\linewidth]{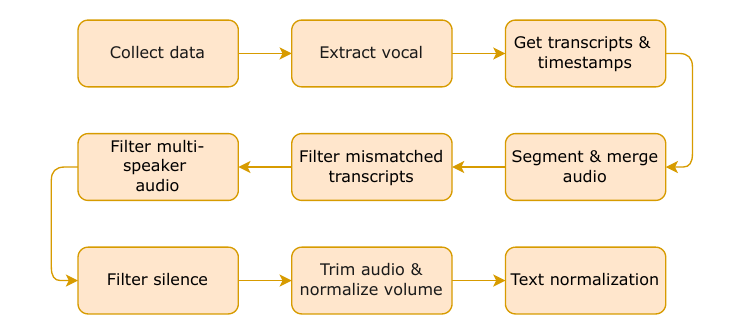}
    \caption{PhoAudiobook creation pipeline.}
    \label{fig:phoaudiobook-pipeline}
\end{figure}

Figure \ref{fig:phoaudiobook-pipeline} illustrates the creation process of our PhoAudiobook dataset. 

First, we collect raw Vietnamese audiobook data from the publicly accessible website \url{https://sachnoiviet.net}. This raw dataset includes 23K hours of content from 2,697 audiobooks, narrated by 735 distinct speakers. Next, we use \texttt{demucs} to extract the vocal track, effectively removing any background music or sound effects \cite{defossez2021hybrid,rouard2022hybrid}. We then employ the multilingual \texttt{Whisper-large-v3} model to generate transcriptions and corresponding timestamps for the audio data \cite{whisper}.
The output from \texttt{Whisper-large-v3} includes transcripts for short audio segments, usually a few seconds long, along with their corresponding timestamps. These segments are often aligned with natural pauses in speech. We then concatenate \textit{successive} audio segments and their corresponding transcripts to create longer audio samples and transcriptions, each lasting between 10 and 20 seconds. 
To ensure the quality of the transcriptions, we process these merged samples using the state-of-the-art Vietnamese ASR model, \texttt{PhoWhisper-large} \cite{PhoWhisper}. We then retain only the samples where the \texttt{Whisper-large-v3}-based transcription matches exactly with the transcription output from \texttt{PhoWhisper-large}. 
Furthermore, we tackle the challenge of multi-speaker audio. We use the \texttt{wav2vec2-bartpho} model to identify and filter out short audio samples containing multiple speakers, ensuring that all audio segments associated with a particular speaker are indeed spoken by that individual.\footnote{\url{https://huggingface.co/nguyenvulebinh/wav2vec2-bartpho}}

To reduce excessive silence in the audio data, we exclude samples with transcripts shorter than 25 words and trim silence from the beginning and end of each sample. Additionally, we use the \texttt{sox} library to normalize audio volume levels for maintaining consistency and avoiding abrupt loudness throughout the dataset.\footnote{\url{https://sourceforge.net/projects/sox}} 
Finally, we standardize the transcriptions through a text normalization step, which includes converting text to lowercase, adding appropriate punctuation, and normalizing numerical expressions into their text form (e.g., "43" becomes "forty three"). We carry out this text normalization step using a sequence-to-sequence model, which we develop by fine-tuning the pre-trained \texttt{mbart-large-50} model \cite{liu-etal-2020-multilingual-denoising} on a Vietnamese dataset consisting of unnormalized input and normalized output text pairs.\footnote{\url{https://huggingface.co/datasets/nguyenvulebinh/spoken_norm_pattern}}

The data creation process described above results in a refined 1,400-hour audio corpus. To ensure balanced speaker representation, we limit each speaker to a maximum of 4 hours of audio. This results in a high-quality dataset comprising \textbf{941} hours of audio from \textbf{735} speakers. From the remaining $1,400 - 941 = 559$ hours of audio, we sample 0.8 hour of audio from 20 speakers to construct a "seen" speaker test set. Additionally, we split the 941 hours of audio from 735 speakers into three sets (on speaker level): a training set containing 940 hours from 710 speakers, a validation set with 0.5 hours from 5 speakers, and an "unseen" speaker test set comprising 0.4 hours from 20 speakers who have the shortest total audio durations. 
Here, the 20 speakers in the "seen" speaker test set are part of the 710 speakers used for training.

We conduct a post-processing step to manually inspect each audio sample and its corresponding transcription from both the "seen" and "unseen" speaker test sets. This process results in all correct transcriptions in the test sets of PhoAudiobook.

\subsection{Dataset analysis}
Table~\ref{tab:dataset_stats} presents the characteristics of our dataset -- PhoAudiobook, in comparison to other Vietnamese speech datasets, including VinBigData~\cite{vinbigdata2023}, VietnamCeleb~\cite{vietnamceleb}, the VLSP 2020 ASR Challenge,\footnote{\url{https://vlsp.org.vn/vlsp2020/eval/asr}} BUD500~\cite{bud500}, and viVoice~\cite{vivoice}.

\begin{table*}[t]
    \centering
    \renewcommand{\arraystretch}{1.2}
    \resizebox{\textwidth}{!}{
    \begin{tabular}{lccccccccc}
    \toprule
    \multicolumn{1}{l}{\textbf{Dataset}} & \multicolumn{1}{c}{\shortstack{\textbf{Duration} \\ (h)}} & \multicolumn{1}{c}{\shortstack{\textbf{Mean Dur.} \\ (s)}} & \multicolumn{1}{c}{\shortstack{\textbf{25\% Dur.} \\ (s)}} & \multicolumn{1}{c}{\shortstack{\textbf{75\% Dur.} \\ (s)}} & \multicolumn{1}{c}{\textbf{Domain}} & \multicolumn{1}{c}{\shortstack{\textbf{SI-SNR} \\ (dB)}} & \multicolumn{1}{c}{\textbf{\# Speakers}} & \multicolumn{1}{c}{\shortstack{\textbf{Rate} \\ (wpm)}} & \multicolumn{1}{c}{\shortstack{\textbf{Fs} \\ (Hz)}} \\
    \midrule
    VinBigData  & 101  & 6.47  & 3.54  & 8.09 & General-purpose & 4.77 & Unknown & 229 & 16000 \\
    VietnamCeleb & 187 & 7.74 & 2.84 & 9.64 & Unknown & 3.89 & Unknown & No transcripts & 16000 \\
    VLSP & 243 & 4.37 & 2.43 & 5.21 & Unknown & 4.35 & Unknown & 242 & 16000 \\
    BUD500 & 462 & 2.56 & 2.11 & 2.94 & General-purpose & 4.22 & Unknown & 224 & 16000 \\
    viVoice & 1016 & 4.12 & 1.96 & 5.55 & General-purpose & 4.81 & Unknown & 243 & 24000 \\
    PhoAudiobook & 941 & 11.66 & 10.63 & 12.18 & Audiobooks & 4.91 & 735 & 201 & 16000 \\
    \bottomrule
    \end{tabular}
    }
    \caption{Characteristics of PhoAudiobook and other speech datasets for Vietnamese.}\label{tab:dataset_stats}
\end{table*}

\begin{figure}[t]
    \centering
    \includegraphics[width=1\linewidth]{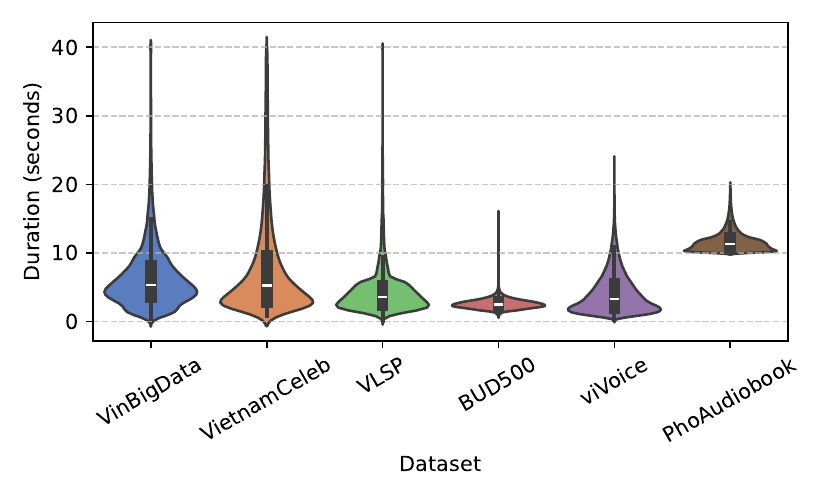}
    \caption{Duration distributions of datasets. Audio samples are capped at 40 seconds for visualization purposes.}\label{fig:dur-dist}
\end{figure}

\paragraph{Duration:} PhoAudiobook, with 941 hours, is the second-largest dataset, closely following viVoice, which has 1,016 hours. The other datasets are considerably smaller, ranging from 101 to 462 hours. Figure \ref{fig:dur-dist} shows that previous datasets primarily consist of audio segments shorter than 10 seconds. PhoAudiobook addresses this limitation by providing audio samples ranging from 10 to 20 seconds.

\paragraph{Domain:} PhoAudiobook is derived from audiobooks, typically recorded with professional equipment in controlled environments, ensuring high-quality audio. In contrast, other datasets are general-purpose (e.g., news, YouTube videos, conversations) and may include audio recorded on consumer devices in uncontrolled settings, often with background noise. However, general-purpose datasets have the advantage of covering diverse topics and speaking styles.

\paragraph{Signal-to-Noise Ratio (SI-SNR):} Using an SI-SNR estimator from the speechbrain toolkit \cite{speechbrain}, we calculated SI-SNR across 1000 randomly sampled audio from each dataset. PhoAudiobook achieves the highest SI-SNR, surpassing all other datasets, including viVoice.

\paragraph{Speaker Information:} PhoAudiobook is the only dataset with explicit speaker identity, and therefore, number of speakers (735).

\paragraph{Speaking Rate (wpm):} Among the four datasets with transcripts, PhoAudiobook has the lowest speaking rate of words per minute. This reflects the nature of the dataset, which features long-form audio where speakers naturally pause and rest.

\paragraph{Sampling Rate:} All datasets except viVoice use a standard sampling rate of 16000 Hz, which is a widely used sampling rate for speech data.\\

PhoAudiobook is comparable in total duration to viVoice, however, it offers several advantages:

\begin{itemize}[leftmargin=*]
\setlength\itemsep{-1pt}

    \item \textbf{Text Normalization:} viVoice lacks text normalization, which limits its suitability for certain TTS models. In contrast, PhoAudiobook offers normalized transcripts, enhancing compatibility with these models.
    \item \textbf{Audio Quality:} The unnormalized audio waveforms in viVoice may cause quality issues like distortion and inconsistent volume. In contrast, PhoAudiobook ensures audio waveforms are normalized for consistent quality. 
    \item \textbf{Speaker ID:} viVoice does not provide speaker IDs for individual audio samples, but uses YouTube channel names as a proxy. This approach can be problematic when a YouTube channel features multiple speakers, limiting the use of this dataset to models that do not require speaker identification. In contrast, PhoAudiobook provides distinct speaker IDs for each audio sample, ensuring its broader applicability for speaker-dependent tasks. 

\end{itemize}

\section{Empirical approach}
\label{sec:exp}

\subsection{Models \& Training data augmentation} 

We conduct experiments using 3 state-of-the-art zero-shot TTS models: VALL-E \cite{valle}, VoiceCraft \cite{voicecraft}, and XTTS-v2 \cite{xtts}. (i) VALL-E, a pioneering language model-based approach, treats text-to-speech (TTS) as a conditional language modeling task. It utilizes discrete acoustic tokens derived from a neural audio codec and leverages massive datasets to achieve impressive zero-shot, in-context learning capabilities.
(ii) VoiceCraft, a token-infilling neural codec language model, excels in both speech editing and zero-shot TTS. It employs a Transformer decoder architecture with a novel token rearrangement procedure to generate high-quality speech.
(iii) XTTS-v2 builds upon the Tortoise model \cite{betker2023betterspeechsynthesisscaling}, incorporating modifications for multilingual training and enhanced voice cloning. It excels in synthesizing speech for numerous languages, including low-resource ones.

To enhance data distribution and ensure our TTS models effectively handle shorter input text, we augment the PhoAudiobook training set with shorter audio clips. Specifically, we treat the PhoAudiobook training set, which consists of 940 hours, as a new raw dataset and apply our dataset creation process as detailed in Section \ref{subsec:phoaudiobook}. However, we omit (i) the step of merging short segments into longer ones and (ii) the step of excluding short samples. This augmentation phase results in an additional 554 hours of short audio, bringing the total to 940 + 554 = 1494 hours of audio for training. 
See implementation details on how we train VALL-E, VoiceCraft, and XTTS-v2 on this 1494-hour training set in Appendix~\ref{app:implementation}.

\subsection{Evaluation setup}

\paragraph{Baseline:} The baseline model is viXTTS \cite{vivoice},\footnote{\url{https://huggingface.co/capleaf/viXTTS}} which is fine-tuned from the pre-trained XTTS-v2 on the viVoice dataset.

\paragraph{Test sets:}  In addition to using our PhoAudiobook "seen" and "unseen" speaker test sets, we also compare our models with viXTTS on the VIVOS test set \cite{vivos}, which contains 0.75 hours of short audio data from 19 speakers. Furthermore, we randomly select 8 speakers, totaling 0.5 hours of audio, from the viVoice dataset for testing. It is important to note that viVoice is available only as a single training dataset, without a predefined training/validation/test split. Consequently, this 0.5-hour viVoice audio set is in fact used for training the baseline viXTTS.

\paragraph{Metrics:} To compare our models and the baseline, we use objective metrics including Word Error Rate (WER), Mel-Cepstral Distortion (MCD) and F0 Root Mean Square Error (RMSE$_{F0}$), as well as subjective metrics Mean Opinion Score (MOS)  and Similarity MOS (SMOS). 

\textbf{Objective metrics} provide quantifiable measures of specific aspects of synthesized speech:

\begin{itemize} 
\item \textbf{Word Error Rate (WER):} This metric assesses the intelligibility of synthesized speech by calculating the edit distance between the transcription of the synthesized speech and the ground truth transcription. Specifically, it counts the number of insertions, deletions, and substitutions needed to turn one into the other. A lower WER indicates higher intelligibility. Here, we employ the ASR  model \texttt{PhoWhisper-large}~\cite{PhoWhisper} to generate the transcription of the synthesized speech.

\item \textbf{Mel-Cepstral Distortion (MCD):} This metric quantifies the spectral difference between synthesized speech and the ground truth audio. A lower MCD value indicates higher spectral similarity and better quality. We use the \texttt{pymcd}\footnote{\url{https://pypi.org/project/pymcd/}} package to compute the MCD.

\item \textbf{F0 Root Mean Square Error (RMSE$_{F0}$):} This metric measures the difference in fundamental frequency (F0) between synthesized speech and the ground truth audio. A lower RMSE$_{F0}$ suggests a better matching of intonation and prosody. We use the \texttt{Amphion}~\cite{amphion} toolkit to compute this value.
\end{itemize}

\textbf{Subjective Metrics} rely on human judgments to evaluate the overall quality and naturalness of synthesized speech.

\begin{itemize} 
\item \textbf{Mean Opinion Score (MOS):} This metric assesses the overall quality of synthesized speech, taking into account factors such as naturalness, clarity, and listening effort. Human listeners rate the speech on a scale from 1 (very poor) to 5 (excellent).

\item \textbf{Similarity MOS (SMOS):} This metric evaluates the perceived speaker similarity between the speech prompt and the generated speech. Listeners rate the similarity on a scale from 1 (completely different) to 5 (identical).

\end{itemize}

To conduct the subjective evaluation, we first randomly sample one audio file from each speaker in the test set. We then hire 10 native speakers to rate the outputs for the in-distribution test sets (PAB-S, PAB-U) and 20 native speakers for the out-of-distribution test sets (VIVOS, viVoice), with all ratings on a scale from 1 to 5, using 0.5-point increments. To ensure fairness, we shuffle and anonymize the model names so that each listener is unaware of which model produces each sample.

\begin{table}[!t]
\centering
\resizebox{7.5cm}{!}{
\setlength{\tabcolsep}{0.3em}
\def\arraystretch{1.1}
\begin{tabular}{l|l | r | r | r | r }
\hline
\multicolumn{2}{c|}{\textbf{Model}} & \textbf{PAB-S} & \textbf{PAB-U} & \textbf{VIVOS} & \textbf{viVoice} \\
\hline
\multirow{5}{*}{\rotatebox[origin=c]{90}{\textbf{WER $\downarrow$}}} 
& Original & 0.88 & 0.83 & 5.14 & 4.97 \\
\cdashline{2-6}
& VALL-E\textsubscript{PAB} & 24.96 & 12.90 & \textbf{12.63} & 13.58 \\
& VoiceCraft\textsubscript{PAB} & 7.53 & 15.14 & \underline{13.53} & 21.70\\
& XTTS-v2\textsubscript{PAB} & \textbf{4.16} & \textbf{4.31}& {37.81}	&\textbf{8.32}\\
\cdashline{2-6}
& viXTTS & \underline{4.23}	& \underline{5.17} & {37.81}	& \underline{12.54} \\
\hline
\multirow{4}{*}{\rotatebox[origin=c]{90}{\textbf{MCD $\downarrow$}}} & VALL-E\textsubscript{PAB} & 7.50	& 8.28 & \underline{10.13} & \underline{8.70}\\
& VoiceCraft\textsubscript{PAB} & \underline{6.69} & \underline{7.98} & 10.27 & 9.15\\
& XTTS-v2\textsubscript{PAB} & \textbf{6.30} & \textbf{7.81} & \textbf{9.85} & \textbf{8.34} \\
\cdashline{2-6}
& viXTTS & 7.47 & 8.48 & 10.54 & 8.71 \\
\hline
\multirow{4}{*}{\rotatebox[origin=c]{90}{\textbf{RMSE$_{F0}$ $\downarrow$}}} & VALL-E\textsubscript{PAB} & 226.55 & \underline{246.88} & \underline{267.80} & \textbf{223.56}\\
& VoiceCraft\textsubscript{PAB} & \textbf{214.66} & 247.54 & \textbf{259.46} & 233.68 \\
& XTTS-v2\textsubscript{PAB} & \underline{216.44} & \textbf{242.51} & 290.77 & \underline{228.81} \\
\cdashline{2-6}
& viXTTS & 249.54 & 271.70 & 338.59 & 238.05 \\
\hline
\multirow{5}{*}{\rotatebox[origin=c]{90}
{\textbf{MOS $\uparrow$}}} 
& Original & $4.61 \pm 0.17$ & $4.63 \pm 0.16$ & $4.41 \pm 0.14$ & $4.66 \pm 0.20$ \\
\cdashline{2-6}
& VALL-E\textsubscript{PAB} & $3.96 \pm 0.29$ & \textbf{4.04} $\pm\ 0.28$ & \underline{3.44} $\pm\ 0.21$ & $3.75 \pm 0.38$ \\
& VoiceCraft\textsubscript{PAB} & \underline{4.16} $\pm\ 0.21$ & $3.75 \pm 0.29$ & \textbf{3.85} $\pm\ 0.22$ & \textbf{3.98} $\pm\ 0.22$ \\
& XTTS-v2\textsubscript{PAB} & \textbf{4.20} $\pm$ 0.20 & \underline{3.89} $\pm\ 0.21$ & $2.79 \pm 0.21$ & \underline{3.98} $\pm\ 0.29$ \\
\cdashline{2-6}
& viXTTS & $4.05 \pm 0.23$ & $3.85 \pm 0.25$ & $2.37 \pm 0.24$ & $3.48 \pm 0.44$ \\
\hline

\multirow{5}{*}{\rotatebox[origin=c]{90}{\textbf{SMOS $\uparrow$}}}
& Original & $4.23 \pm 0.23$ & $3.90 \pm 0.32$ & $3.87 \pm 0.24$ & $3.34 \pm 0.47$ \\
\cdashline{2-6}
& VALL-E\textsubscript{PAB} & \textbf{3.77} $\pm\ 0.24$ & \underline{3.46} $\pm\ 0.29$ & \textbf{3.35} $\pm\ 0.25$ & $3.20 \pm 0.38$ \\
& VoiceCraft\textsubscript{PAB} & \underline{3.64} $\pm\ 0.30$ & $3.32 \pm 0.35$ & \underline{3.25} $\pm\ 0.25$ & \textbf{3.41} $\pm\ 0.36$ \\
& XTTS-v2\textsubscript{PAB} & $3.55 \pm 0.27$ & \textbf{3.56} $\pm\ 0.29$ & 3.03 $\pm\ 0.23$ & \underline{3.39} $\pm\ 0.41$ \\
\cdashline{2-6}
& viXTTS & $2.88 \pm 0.28$ & $2.63 \pm 0.32$ & $2.48 \pm 0.23$ & $3.11 \pm 0.43$ \\
\hline
\end{tabular}
}
\caption{Test results of different TTS models. Our models, "VALL-E\textsubscript{PAB}", "VoiceCraft\textsubscript{PAB}" and "XTTS-v2\textsubscript{PAB}" are obtained by training VALL-E, VoiceCraft, and XTTS-v2 on our PhoAudiobook training data, respectively. "PAB-S" and "PAB-U" refer to the PhoAudiobook "seen" and "unseen" speaker test sets, respectively. The viXTTS model is fine-tuned from the pre-trained XTTS-v2 using the entire viVoice dataset.}
\label{tab:mainresults}
\end{table}

\section{Results}
\label{sec:results}
Table \ref{tab:mainresults} presents the results obtained for our trained models and the baseline. It is clear that our XTTS-v2\textsubscript{PAB} consistently outperforms viXTTS across all metrics and test sets. For instance, on the viVoice set, XTTS-v2\textsubscript{PAB} achieves the best WER of 8.32, which is substantially lower than the 12.54 WER of viXTTS, even though viXTTS is tested on its own training data. Additionally, XTTS-v2\textsubscript{PAB} also produces substantially higher SMOS and RMSE$_{F0}$ scores compared to viXTTS in all test sets, indicating that the speech it generates more closely resembles the reference speaker. These results suggest that XTTS-v2\textsubscript{PAB} outputs more intelligible and natural-sounding speech that better captures the nuances of the target speaker's voice, for both long (PhoAudiobook and viVoice) and shorter (VIVOS) text inputs.
 
We observe a variation in the performance of different models across test sets. While VoiceCraft\textsubscript{PAB} and VALL-E\textsubscript{PAB} are less competent than XTTS-v2\textsubscript{PAB} on the test sets PAB-S, PAB-U and viVoice, they outperform XTTS-v2\textsubscript{PAB} on the VIVOS test set. Specifically, for PAB-S, PAB-U, and viVoice test sets, VALL-E\textsubscript{PAB} and VoiceCraft\textsubscript{PAB} underperform compared to XTTS-v2\textsubscript{PAB} in terms of WER, while achieving comparable results on other metrics such as MCD, RMSE$_{F0}$, MOS, and SMOS. However, on the VIVOS test set, XTTS-v2\textsubscript{PAB} and viXTTS perform significantly worse than VALL-E\textsubscript{PAB} and VoiceCraft\textsubscript{PAB} across all evaluation metrics. This indicates that VALL-E\textsubscript{PAB} and VoiceCraft\textsubscript{PAB} are more adept at handling short sentences, which are characteristic of the VIVOS test set. Upon manual inspection, we found that for short text inputs, XTTS-v2-based models -- XTTS-v2\textsubscript{PAB} and viXTTS -- often generate redundant or rambling speech at the end of the output. This suggests a potential architectural issue within the XTTS-v2 model itself, rather than a data-related problem, as both the viVoice dataset and the "augmented" PhoAudiobook training set contain short audio samples.

\section{Conclusion}
\label{sec:conclusion}
We have introduced PhoAudiobook, a comprehensive 941-hour high-quality dataset designed for Vietnamese text-to-speech (TTS) synthesis. Using this dataset, we conducted experiments with three leading zero-shot TTS models: VALL-E, VoiceCraft, and XTTS-v2. Our findings show that XTTS-v2 consistently outperforms its counterpart trained on the viVoice dataset across all metrics, highlighting the superiority of PhoAudiobook in enhancing model performance. Additionally, VALL-E and VoiceCraft demonstrate exceptional capability in handling short sentences.

\section*{Limitations}
While models trained on PhoAudiobook show high performance on purely Vietnamese datasets, we have not evaluated their performance in code-switching scenarios where the input text includes both Vietnamese and English. Future research should investigate the models' ability to handle multilingual inputs to enhance their applicability in more diverse linguistic contexts.

\section*{Acknowledgments}
This work was completed while all authors were at Movian AI, Vietnam. All datasets and models were downloaded, trained, and evaluated using Movian AI's resources.

We would like to express our sincere gratitude to  \href{https://nguyennm1024.github.io/}{Mr. Nguyen Nguyen} for providing a toolkit that facilitated the process of downloading raw data from \url{https://sachnoiviet.net}.

\bibliography{refs}

\newpage 

\appendix

\section{Implementation details}\label{app:implementation}

For the VALL-E implementation, the first phase is the data processing stage. We fine-tune a model based on the Vietnamese Wav2Vec 2.0 large model\footnote{\url{https://huggingface.co/nguyenvulebinh/wav2vec2-large-vi}} for the Vietnamese dialect recognition task using our in-house data. The model achieves an accuracy of approximately 95\% for the dialect recognition task. We then apply this model to the PhoAudiobook. To accurately determine the regional dialect for each speaker and reduce inference time, we sample about 20 audios for each speaker and feed them into the model. We assign the dialect of each speaker based on the region with the highest number of predicted audios. Because the VALL-E model is trained based on the phoneme level, we use the "phonemizer" package to convert the text into phonemes based on the dialect of each speaker.\footnote{\url{https://github.com/bootphon/phonemizer}} For the audio, we also use the Audio CodeC encoder to compress raw audio into discrete tokens. 
The second phase is model training. We employ 12 Transformer-decoder layers, each with 1024 hidden units and 16 attention heads. We use a batch size corresponding to a maximum duration of 40 audio seconds, a base learning rate of 0.05, and 4 gradient accumulation steps. Our model is trained on 8 A100-40GB GPUs. Our implementation is based on the customized GitHub repository that reproduces the idea from the VALL-E paper.\footnote{\url{https://github.com/lifeiteng/vall-e/tree/main}} We make modifications to this repository for our specific language and data.

Since VoiceCraft takes phoneme representations as input, we first convert our data to phonemes using the "phonemizer" package. We then append the derived Vietnamese phonemes to the existing English vocabulary and expand the text embedding layer to accommodate Vietnamese phonemes. The 830M\_TTSEnhanced checkpoint is a public VoiceCraft model fine-tuned with a text-to-speech objective and serves as our starting point. Following the author's implementation,\footnote{\url{https://github.com/jasonppy/VoiceCraft}} we fine-tune this model using the AdamW optimizer with a learning rate of $1e^{-5}$ and a batch size of $25,000$ tokens, which corresponds to approximately $8.3$ minutes of audio. We train the model on 4 A100-40GB GPUs for $16$ epochs.

XTTS-v2 employs BPE for text encoding. To adapt the training to Vietnamese data, we use the same Vietnamese token list employed by  \citet{vivoice}. We follow the training recipes provided in the coqui's TTS repository and fine-tune the public XTTS-v2 checkpoint trained for 16 languages.\footnote{\url{https://github.com/coqui-ai/TTS}} We extend the character and audio length limits to accommodate audio segments up to 20 seconds in duration for training data. We use the AdamW optimizer with a learning rate of $5e^{-6}$, a batch size of $4$, and fine-tune the model on a single A100-40GB GPU for $18$ epochs.

For all these 3 models, we select the model checkpoint
that obtains the best loss on the PhoAudiobook validation set. 

\end{document}